\documentclass[10pt]{article} 
\usepackage[preprint]{rlc}

\usepackage{amssymb}            
\usepackage{mathtools}          
\usepackage{mathrsfs}           
\mathtoolsset{showonlyrefs}     
\usepackage{graphicx}           
\usepackage{subcaption}         
\usepackage[space]{grffile}     
\usepackage{url}                
\usepackage{subcaption}
\usepackage{comment}

\title{Monitoring Fidelity of Online Reinforcement Learning Algorithms in Clinical Trials}

\author{Anna L. Trella  \\
    annatrella@g.harvard.edu \\
    School of Engineering and Applied Sciences \\
    Harvard University
    \And
    Kelly W. Zhang  \\
    kelly.w.zhang@columbia.edu \\
    Graduate School of Business \\
    Columbia University
    \And
    Inbal Nahum-Shani  \\
    inbal@umich.edu \\
    Institute for Social Research \\
    University of Michigan
    \And
    Vivek Shetty  \\
    vshetty@ucla.edu \\
    Schools of Dentistry \& Engineering \\
    University of California, Los Angeles
    \And
    Iris Yan  \\
    irisyan@college.harvard.edu \\
    School of Engineering and Applied Sciences \\
    Harvard University
    \And
    Finale Doshi-Velez  \\
    finale@seas.harvard.edu \\
    School of Engineering and Applied Sciences \\
    Harvard University
    \And
    Susan A. Murphy  \\
    samurphy@g.harvard.edu \\
    School of Engineering and Applied Sciences \\
    Harvard University
    }

\begin{document}

\maketitle

\begin{abstract}
Online reinforcement learning (RL) algorithms offer great potential for personalizing treatment for participants in clinical trials. However, deploying an online, autonomous algorithm in the high-stakes healthcare setting makes quality control and data quality especially difficult to achieve. This paper proposes \textit{algorithm fidelity} as a critical requirement for deploying online RL algorithms in clinical trials. It emphasizes the responsibility of the algorithm to (1) safeguard participants and (2) preserve the scientific utility of the data for post-trial analyses. We also present a framework for pre-deployment planning and real-time monitoring to help algorithm developers and clinical researchers ensure algorithm fidelity. To illustrate our framework's practical application, we present real-world examples from the Oralytics clinical trial. Since Spring 2023, this trial successfully deployed an autonomous, online RL algorithm to personalize behavioral interventions for participants at risk for dental disease.
\end{abstract}

\section{Introduction}
\label{intro}
Reinforcement learning (RL) algorithms are increasingly used in digital health clinical trials to determine the delivery of interventions via devices like wearables or mobile apps \citep{DBLP:journals/corr/abs-1909-03539, yom2017encouraging,figueroa2021adaptive, forman2019can}.
Recently, there has been growing interest in the \textit{online} class of RL algorithms that update their policies using incoming participant data throughout the trial. 
This is because online RL algorithms have greater potential to learn more effectively (e.g., handle user heterogeneity, distribution shift, non-stationarity, etc.) and provide personalized interventions. However,  online algorithms require real-time updates and interactions with complex systems. They can result in issues that, if not properly monitored and planned for, jeopardize the entire trial. These issues are especially concerning because clinical trials can be expensive and take years to develop with multiple stakeholders.

Undetected issues or issues solved in an untimely manner can result in (1) insufficient data for conducting post-trial analyses and (2) compromise the effectiveness of the intervention and participant experience. Given the complexity and real-time nature of implementing online RL algorithms in healthcare settings, unexpected issues may arise even with careful planning (e.g., updates with overflow values result in an incorrect policy; state or reward data does not properly write to the database). There are well-established criteria and a variety of existing frameworks for 
real-world machine learning (ML) deployment \citep{oala2021machine,challen2019artificial,loftus2022ideal,sambasivan2021everyone}. While some of these frameworks are specific to health, the inherent properties of online RL algorithms introduce unique challenges that this literature does not address.

We first introduce the critical concept of \textit{algorithm fidelity} in clinical trials (Section~\ref{sec_fidelity_def}), which refers to the responsibility of the online RL algorithm to (1) safeguard participants and (2) preserve the scientific utility of data for post-trial analyses. Then, we provide a framework for helping research teams ensure algorithm fidelity through proper planning and building a monitoring system around any online RL algorithm (Section~\ref{sec_framework}). Our work brings to the RL community's attention the critical gap between fast-paced algorithmic research and the feasibility of deploying these algorithms effectively (i.e., in a way that does more good than harm)
in high-stakes clinical trials. To showcase the utility of the algorithm fidelity framework, we share insights and issues identified by the real-time monitoring system built for Oralytics. Oralytics successfully deployed an online RL algorithm in an oral health clinical trial that has been in the field since Spring 2023.

\section{Running Example: Supporting Oral Health}
To facilitate the use of our algorithm fidelity framework (Section~\ref{sec_framework}), we show concrete examples of how we used our framework to plan and develop an autonomous monitoring system for a real RL algorithm involved in an oral health clinical trial. As our running example, we use Oralytics \citep{oralytics:clinicaltrial, a15080255, trella2023reward, nahum2024optimizing}: a digital behavioral intervention trial with participants at risk for dental disease. The trial involved an online RL algorithm to optimize the delivery of prompts to encourage participants to engage in oral self-care behaviors. 

Each trial participant received a commercially available electric toothbrush with Bluetooth connectivity and built-in sensors and was asked to download the Oralytics mobile application onto their smartphone. The RL algorithm determined the delivery of prompts (e.g., brushing feedback and motivational messages) via push notifications from the Oralytics app. In Spring 2023, a pilot version of the RL algorithm was deployed for Oralytics with a cohort of 10 participants. In Fall 2023, the finalized RL algorithm was deployed in the main clinical trial of Oralytics with 70 participants, recruited incrementally with around 5 participants starting in the trial every 2 weeks. The trial duration for each participant in the main trial was 70 days. Every day, the RL algorithm determined the delivery of prompts for each participant. There were two decision times when a prompt may be delivered: (1) one hour before the participant-specified morning brushing time, and (2) one hour before the participants-specified evening brushing time. The RL algorithm updated weekly on Sundays with the entire history of data across all participants and up to that time.

\subsection{Defining the RL Problem}
\label{rl_basics}

We review the definitions of state, action, and reward. For each of the components, we use subscript $i \in [1 \colon N]$ to denote the participant and subscript $t \in [1 \colon T]$ to denote the decision time. In addition, online RL algorithms can be decomposed into two main parts: 1. a model of the environment (participant behavior) and 2. a policy informed by the environment model. The RL algorithm fits the model of the environment and iteratively updates that model using a history of states, actions, and rewards from participants. In Oralytics, the RL algorithm was a generalized contextual bandit that performed Thompson sampling \citep{russo2018tutorial}. Therefore, the model of the environment was a model of the expected reward and the policy was an action-selection probability formed using a posterior distribution.

The state $S_{i, t}$ for participant $i$ at decision time $t$ consists of relevant features of the participant environment 
that provide the algorithm with the current context of the participant. The state features could include information such as current weather, the participant's current location, or recent adherence to the  trial protocol. In Oralytics, the state features included the time of day, how well a participant brushed in the past week, amount of prompts sent to the participant (dosage) in the past week, and recent app engagement. Notice that while some features are deterministically calculated (i.e., time of day) other features require current sensor information from the toothbrush and app. See Appendix~\ref{app:alg_state} for a full discussion of Oralytics state features.

An action $A_{i, t}$ (i.e., intervention) is decided by the algorithm for participant $i$ at decision time $t$. For Oralytics, the actions $A_{i,t} \in \{0, 1\}$ were binary and referred to whether to send a prompt $(A_{i,t} = 1)$ or not send one $(A_{i,t} = 0)$. To select actions, the algorithm observed state $S_{i, t}$, formed an action-selection probability $\pi_{i,t}$, and selected $A_{i,t} = 1$ with that probability.
The RL algorithm in Oralytics was a Thompson sampling-based algorithm so 
this action-selection probability $\pi_{i, t}$ was formed using
the posterior distribution maintained by the algorithm. See Appendix~\ref{app:action_space} for more details on the Oralytics action space and policy.

The reward $R_{i, t}$ is a function of the participant's health outcome following the selection of $A_{i,t}$.
The goal of the algorithm is to maximize the total expected rewards collected across participants over the entire trial, $\mathbb{E} \left[ \sum_{t = 1}^T \sum_{i = 1}^N R_{i, t} \right]$. In Oralytics, the reward was a function of brushing quality (proximal health outcome) and a proxy for participant burden (cost of current action on the effectiveness of future actions) \citep{trella2023reward}. See Appendix~\ref{app:reward} for more details on this reward.


\begin{figure}[t]
    \centering
    \begin{subfigure}{0.8\textwidth}
        \includegraphics[trim={1.1cm 1.5cm 1.4cm 3.8cm},clip,width=1\textwidth]{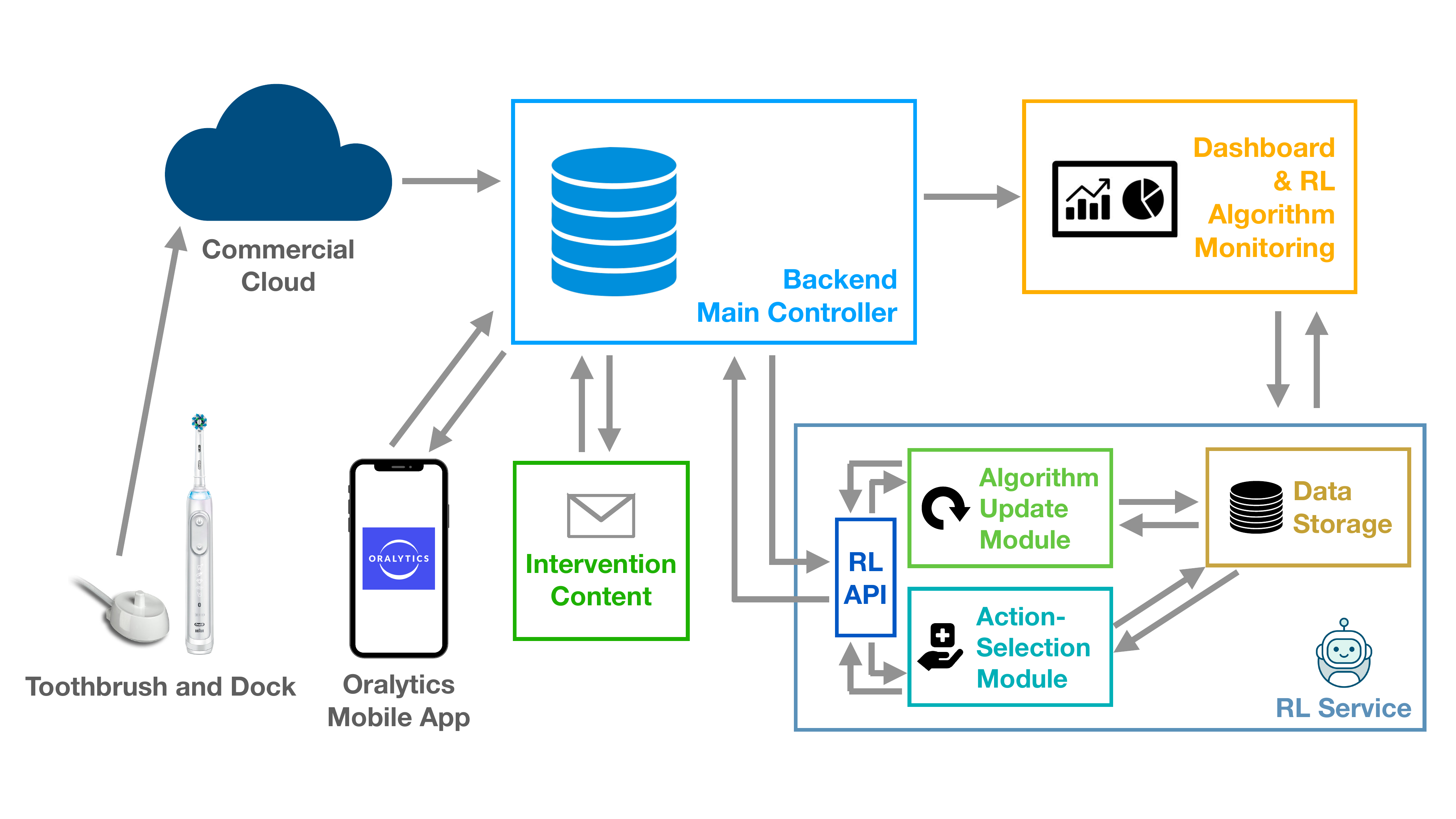}
    \end{subfigure}
    \caption{Oralytics System and RL System Architecture. Brushing data is captured by sensors in the toothbrush and uploaded to the commercial cloud via a dock. The main controller gathers this data, along with app engagement data from the Oralytics app, and feeds it to the RL service and the dashboards. This sensor data is provided to the RL service to select actions and update. Using the actions selected by the RL service, the main controller populates intervention prompt content, and schedules prompts onto each participant's Oralytics app.}
    \label{fig:oralytics_architecture}
\end{figure}

\subsection{Oralytics System Components}
\label{sec_system_components}
The RL algorithm is one component of the RL software service. The RL service consists of (1) an API to communicate with other services, (2) a dependency module to request sensor data from other services, (3) a module to update the RL algorithm based on newly collected participant data, (4) an action-selection module that takes in the participant's current state and assigns an intervention based on the algorithm's current policy, and (5) an internal database that stores all the data necessary for the RL algorithm to function and perform post-trial analyses. 

To function properly, the RL service depends on stable interactions with a variety of other services. We refer to this larger ecosystem of software components as the Oralytics system (Figure~\ref{fig:oralytics_architecture}). When a participant brushed, the toothbrush recorded brushing sensor data from that session and when docked properly, this data was sent to the commercial cloud. Similarly, when a participant uses the Oralytics app, any app analytics data collected by the Oralytics app was sent to the backend main controller. For each participant's decision-time, the backend main controller collected sensor data, made this data available for the RL service to request, and called the RL API's action selection module. Once called, the RL service obtained the recent sensor data, constructed the participants' state features, then returned the actions selected by the RL algorithm's policy. The RL service saved the necessary data corresponding to the state and action to the internal database.
The backend main controller then scheduled prompts accordingly. During algorithm update time, the backend main controller collected sensor data needed for reward construction from the commercial cloud, made this data available for the RL service to request, and called the RL API's algorithm update module. The RL service obtained the recent sensor data, constructed rewards for each new decision-time, and then updated its policy using the entire history of state, action, and reward data. The RL service saved the necessary data corresponding to reward and policy to the internal database. As one can see, proper RL algorithm functionality requires frequent real-time communication between a variety of components. While the trial is running, many things could go wrong.


\section{Defining Algorithm Fidelity}
\label{sec_fidelity_def}
To help research teams best prepare for RL algorithm deployment in the challenging healthcare setting, we introduce \textit{algorithm fidelity}: the responsibility of the online RL algorithm in (1) safeguarding participants and (2) preserving the scientific utility of the data in clinical trials. These concepts build upon quality control and data quality, which are two well-defined criteria for ML deployment. Quality control, used widely throughout engineering, encompasses a comprehensive set of standards or predefined set of requirements that a product must meet before launching. Data quality involves best practices of handling or reporting the data used to train models. Here, we extend quality control and data quality to concerns involving online RL algorithms in clinical trials.
Often times, other system components (see Appendix~\ref{app_other_forms_fidelity}) are also responsible for ensuring quality control and data quality. However, algorithm fidelity specifically refers to areas that are directly controlled or caused by the RL algorithm. For example, monitoring for app crashes and failing to collect app analytics falls under the purview of the clinician dashboard. However, detecting errors in saving state values used to select actions would come under monitoring of algorithm fidelity.

\subsection{Related Work}
\label{related_work}
We build on quality control and data quality discussed in the existing literature on best-practice frameworks for deploying ML models in general and discuss how they can be adapted to online RL algorithms in clinical trials. 
Many of the ML frameworks for ensuring quality control or data quality are in the offline or in batch settings where the algorithm is trained on previous batch data and the model is fixed before deployment.
In these settings, quality control standards could include achieving baseline prediction accuracy \citep{oala2021machine}, handling distribution shifts and reporting levels of confidence \citep{challen2019artificial}, explainability, fairness, and reproducibility \citep{oala2020ml4h, loftus2022ideal}, or preparing for accidents and unintended risks \citep{amodei2016concrete}.
Data quality could include ensuring that models trained on the data can accurately represent and predict for their task \citep{sambasivan2021everyone}, or forming a checklist of what to report about AI algorithms in clinical trials such as which version of the algorithm was used, how input data was acquired, and how the sample size was determined \citep{liu2020reporting}.

\subsection{Quality Control}
\label{quality_control}
The online nature and inherent properties of RL algorithms require additional aspects of quality control. Unlike prediction models, validating the quality of RL algorithms goes beyond straightforward performance metrics like prediction accuracy. Also, online algorithms autonomously update using incoming data from participants in the trial. Previous work covers monitoring the deployment of between-trial algorithm updates with a fixed policy \citep{oala2020ml4h}, but online settings require identifying and addressing issues in real-time of a policy that is frequently updated throughout the trial. Furthermore, validation in the online setting is more complicated than in the offline setting. In the offline setting, training can involve a human-in-the-loop \citep{wu2022survey} for performing hyperparameter search, checking for convergence, or performing clinical evaluations \citep{oala2020ml4h}. An online algorithm, however, can only be validated once the trial has completed (e.g., via off-policy analyses, one can evaluate whether an action  was effective in certain states for a participant).

Considering our clinical trial setting, we also extend quality control to include standards that involve participant experience and safety. In any clinical trial, participant protection from harm and adverse experiences is the most important consideration. Although there is a list of ethical guidelines that the clinical research team must follow (i.e., overseen by Institutional Review Boards \citep{grady2015institutional}) 
and guidelines that must be followed to  register the trial (i.e., at clinicaltrials.gov) \footnote{The framework considers participant protection responsible by the RL algorithm as opposed to general participant protection guidelines in the trial. Note that ethics are priority consideration prior to algorithmic development. For example, in trials where the algorithm actions may have negative side effects (e.g., administering a drug), the action space is restricted by the current state of the participant. 
}, 
unintended harm can occur via the inappropriate choice of actions by the RL algorithm. For example, the algorithm could deliver too many interventions that take the participant away from normative activities and may lead to burden, habituation, and subsequent dropout. On the other hand, the algorithm could deliver too few interventions or none at all, thus diminishing the intended health benefits of the intervention actions. Furthermore, the algorithm could reduce the functionality of the participant's smartphone by depleting the battery or taking up a large amount of storage space.   



\subsection{Data Quality}
\label{sec_scientific_utility}
Achieving data quality for online RL algorithms is also not straightforward. RL algorithms use and store additional types of data as compared to ML algorithms, such as the action-selection probability $\pi_{i, t}$ for every participant $i$ at decision-time $t$. 
Online algorithms also require extra considerations for data reporting and reproducibility. Unlike offline settings where previous work suggests stating the version of the model used \citep{liu2020reporting}, frequent updates of the RL algorithm in the online setting means needing to index and save each policy update, the data that was used in each update, and code versions and software packages if a code change needed to be made during the trial.

We also extend data quality in our setting to focus on preserving the scientific utility of data for post-trial analyses. Post-trial analyses refers to the broad range of analyses conducted after the trial is completed. These kinds of analyses can (1) improve the design of the intervention for future iterations; (2) inform policy and decisions for rolling out the intervention more broadly; and (3) support the development and refinement of health behavior theories. We now highlight some important analyses of interest. At a minimum, the clinical trial must collect data to perform a primary analysis \citep{furberg2012approaches,qian2022microrandomized}, a statistical analysis that addresses the foremost scientific question of the trial. For example, the primary analysis for Oralytics concerns the causal excursion
effect \citep{boruvka2018assessing} of sending versus not sending a prompt on the participant's proximal oral self-care behaviors. Secondary analysis may focus on investigating whether the benefits of sending (versus not sending) a prompt vary by the participant's baseline characteristics (e.g., age and gender), as well as their current state (e.g., recent brushing quality or app engagement). Additional analyses such as off-policy evaluation \citep{thomas2016data, levine2020offline}, other types of causal inference \citep{ hadad2021confidence}, 
and continual learning \citep{parisi2019continual, khetarpal2022towards} 
assess the benefit of using such an algorithm and improve the design of future versions of the RL algorithm. Furthermore, sharing and publishing high-quality trial data helps the broader scientific community design their own algorithm for future trials.

\section{A Pragmatic Framework for Ensuring Algorithm Fidelity}
\label{sec_framework}

To maintain algorithm fidelity, we offer guidelines for both a pre-trial planning phase (Section~\ref{sec:planning_phase}) 
and a real-time monitoring phase (Section~\ref{sec:monitoring}) while the algorithm is running during the trial. We view our framework as a necessary step in the design and pre-deployment process when developing the RL algorithm. Our framework helps research teams determine possible errors and prioritize issues, developing a monitoring framework that ensures a stable system for participant experience. In addition, our framework also helps research teams solidify a data collection procedure that ensures the validity of trial results and for the data to be usable to stakeholders and other research teams after the trial.

\subsection{Planning Phase}
\label{sec:planning_phase}
If you fail to plan, you plan to fail. Many issues that arise during the trial can be avoided with proper planning. We will see in the later section (Section~\ref{sec:monitoring}) that planning can also reduce the severity of certain issues. Here we examine several topics related to quality control and data quality to consider planning for.
\paragraph{Implementing Fallback Methods (Quality Control)}
Ensuring that participants have consistent functionality throughout the trial is an important part of ensuring a quality participant experience. Even if worst-case issues arise (i.e., software system fails or sensor data is unavailable for state and reward construction), participants should still get some intervention and the algorithm should not update with incorrect or corrupted data. To guarantee a baseline level of functionality, the RL algorithm can be paired with fallback methods. The fallback method is a procedure that is executed in lieu of the standard action-selection or update procedure by the RL algorithm. For action selection, it is critical that each participant receives at least some intervention throughout the trial, even if the intervention is not personalized. For Oralytics, this manifests in two ways. First, instead of providing a single action for the current decision time, the Oralytics RL system constructs and provides a full schedule of actions for the entire full trial length (i.e., 70 days for Oralytics) starting at that decision time. For the first day in the schedule ($t' = t, t + 1$), the algorithm uses the current state $S_{i, t}, S_{i, t + 1}$ as input to the action-selection probability $\pi_{i, t}$. For decision times within the first 2 weeks from the first day, the algorithm will use a modified state space instead (See Appendix~\ref{app:modified_rl_features}). Finally, for all decision times afterward, the algorithm selects actions with fixed probability $\pi_{i, t} = 0.5$. 
Second, if the main controller requests the schedule of actions from the RL service and any step in the RL action-selection procedure fails, a schedule of actions with fixed uniform probability $\pi_{i, t} = 0.5$ is deployed. For updating, if issues arise such as malformed data or if recent data is not readily available, then the algorithm stores the data point but does not update its policy with that data point.

\paragraph{Setting Restrictions (Quality Control and Data Quality)}
Setting proper restrictions or constraints is necessary for safeguarding participants and preserving the scientific utility of the data for post-trial analyses. Here, the restrictions designed to safeguard participants are ones the RL algorithm must adhere to and are not the system restrictions specified in the trial protocol (See footnote in Section~\ref{quality_control}). For Oralytics, one of these restrictions for participant protection included specifying thresholds for too many or too few interventions. The number of interventions a participant received was available on the clinician dashboard and was checked by the staff every day. If a staff member noticed that the amount of interventions exceeded the upper and lower thresholds, then they would contact the RL software engineering team. To facilitate post-trial analyses, other restrictions set in Oralytics included purposely deploying a stochastic RL policy to enable continual learning, and restricting the stochastic RL policy to select actions within a pre-specified range (bounded away from $0$ and $1$) to enhance the ability to address scientific questions with sufficient power \citep{yao2021power} (See Appendix~\ref{app:action_space} for further details). 

\paragraph{Collecting Critical Data (Data Quality)}
Recall from Section~\ref{sec_scientific_utility} that one of the main goals of algorithm fidelity is maintaining the scientific utility of the data.
Some analyses require reproducing the algorithm (i.e., re-running or recreating the decisions made by the RL algorithm), which is also important for validating results and ensuring confidence in trial findings within the scientific community.
Digital intervention clinical trials require the coordination of many complex systems (i.e., RL system, smartphones and/or wearables, data storage systems, etc.) and some types of data cannot be recovered if not explicitly saved. Although some data values could be reconstructed after the trial concludes, unexpected issues could cause reconstruction to be difficult (e.g., bugs or code failures due to OS updates may require code changes during the course of the trial, which could make it difficult to reconstruct action selection probabilities).
Therefore, being able to run post-trial analyses hinges on how well the scientific team set up the data collection and storage plan. This data collection plan includes the variety of data fields that need to be collected and plans regarding how to handle/denote missing and delayed data. For Oralytics, the primary analysis involves brushing quality, which is a component of the reward, but not exactly the reward itself. Therefore, the RL algorithm saved all possible data and raw sensory information used in the construction of the reward (See Appendix~\ref{app:reward}). For reproducibility, Oralytics saved the most basic data fields such as states, actions, and rewards as well as action-selection probabilities used by the RL algorithm, seeds for various random processes, code changes and versions of the algorithm. 

\subsection{Real-Time Monitoring Phase}
\label{sec:monitoring}
Autonomously monitoring an online RL algorithm during the trial is important for detecting, alerting, and preventing failures. Failures and errors could arise during the trial that leads to critical situations (e.g., participants did not get the correct intervention or database failed to save important data fields). A quality monitoring system prevents incidents and detects problems as soon as they arise. This allows the research team to immediately identify, triage, and solve the issue which (1) minimizes the time and negative impact on participants and (2) reduces the burden off of the staff in manually monitoring the clinical trial system. 
To set up the monitoring system, research teams should first specify and assign priority to potential issues that could occur during the trial. Once these issues are specified, one can use built-in packages (e.g., exception handling and Flask email for Python) to send automatic emails when an issue occurs. To assist with this specification and prioritization, we suggest categorizing issues into one of three levels of severity: red, yellow, and green. For each category of issues, we also provide a concrete example of how considering this issue helped with the development of Oralytics.




\paragraph{Red Severity (Compromises Algorithm Fidelity)} Red severity issues are the most severe issues that compromise participant experience or the scientific utility of the data and require immediate attention. Red severity issues result in an immediate email to the RL software engineering team and appear on the clinician dashboard.
Participant experience can be compromised if they receive an unreasonable amount of prompts. This could be caused by communication errors between system components (RL algorithm, main controller, app), out-of-range action-selection probabilities due to numerical instability in the RL algorithm, or failure to recognize that a participant has started in the trial. The scientific utility of the data could be compromised if the RL internal data storage fails to properly write values due to integration issues with the database management system. These issues could include failed/timed out/closed connection with the database and/or the database exceeded storage. Examples of red severity checks in Oralytics included checking in the code if the RL algorithm selected an unreasonable amount of prompts (less than the minimum amount or more than the maximum amount) for a participant or checking for failures in the RL internal database saving values. As of Spring 2024, there have been no red severity issues in Oralytics that were caused by the RL algorithm.
\paragraph{Yellow Severity (Compromises RL Algorithm Functionality)}
Yellow severity issues are issues that compromise the ability of the RL algorithm to learn or personalize interventions for participants. Yellow severity issues also result in an immediate email and appear on the monitoring dashboard. However, the urgency to fix a yellow severity issue is lower, allowing more time for the RL software engineering team to resolve. Notice that issues impacting RL algorithm learning and personalizing are not red severity issues because of fallback methods (Section~\ref{sec:planning_phase}). 
RL algorithm learning and personalizing can be compromised if the algorithm (1) does not obtain sensor data in a stable or timely manner and (2) fails to fetch necessary data from the RL internal database. 
In Oralytics, examples of yellow severity checks included checking if requests to obtain sensor data failed, if the request succeeded but there were issues parsing the data, and if there were issues connecting to and fetching data from the RL database. During the Oralytics trial, the monitoring system detected the yellow severity issue of the RL algorithm failing to construct the current state for a participant. This issue involved the RL algorithm failing to obtain app analytics data for a participant, which is part of the state space. The monitoring system automatically detected this issue through the check for issues parsing the sensor information and sent an email to the RL software engineering team. The team triaged and discovered that the max limit of requests per minute to the main controller was exceeded when more participants entered the trial. Exceeding this limit causes the returned response to be empty and therefore un-parseable. The team fixed the issue by coordinating a more efficient request strategy with the main controller team.
\paragraph{Green Severity (Impacts Post-Trial Data Analysis)}
Green severity issues are problems that need to be documented so that post-trial analyses can be adjusted. If not properly accounted for, these issues could compromise the validity of the primary analysis or the opportunity to improve the system for future trials. Green severity issues are only problems if not documented. Handling these issues could involve a list of items to investigate after the trial is over (e.g., consistency in data saved by the main controller with data in the RL internal database). Sometimes handling these issues involves components outside of the RL system (e.g., main controller or app). Therefore, we suggest making contracts with other component teams prior to the trial to agree on who will document what data and which issues.
In Oralytics, some documented items included all issues that arose in the RL service and its dependencies, timestamps and any associated data when the RL algorithm successfully updated and conversely when it did not, and timestamps when the main controller called the RL API and the attempted call has failed. During the Oralytics trial, a green severity issue was detected by clinical staff when the dashboards showed that only a small fraction of participants had prompts scheduled on their apps. As part of the contract between system components, the staff contacted both the main controller and the RL software engineering teams. Both teams triaged and discovered that the RL API was functioning properly, but the main controller was timing out (i.e., did not wait long enough to get the full schedule for every participant currently in the trial) when calling the RL API. Therefore, the main controller incorrectly pushed a blank schedule of prompts for certain participants. The main controller team fixed the issue the next business day and the RL team documented this issue, the participants affected, and the decision times that were affected. 


\section{Discussion}
In this paper, we discuss adherence to algorithm fidelity, an important problem that research teams will inevitably face when implementing an online RL algorithm in a digital intervention clinical trial. Due to the unpredictable nature of the real-world environment and the dynamic nature of online RL algorithms, maintaining algorithm fidelity can become difficult and expensive unless planned for. We used our successful trial to ground pragmatic solutions from our framework for pre-trial planning and building an autonomous monitoring system. While the focus of this work is on clinical trials, we expect many ideas in this paper to be transferable to general digital experimentation (e.g., online education, ads, UI testing, etc.). By offering the algorithm fidelity definition and a framework for achieving algorithm fidelity, we hope to smooth the transition from algorithm development to real-world application: giving other researchers the confidence to deploy their innovative online RL algorithms in real life.



\subsubsection*{Acknowledgments}
\label{sec:ack}
This research was funded by NIH grants IUG3DE028723, P50DA054039, P41EB028242, U01CA229437, UH3DE028723, and R01MH123804. SAM  holds concurrent appointments  at Harvard University and as an Amazon Scholar. This paper describes work performed at Harvard University and is not associated with Amazon.

\bibliography{main}
\bibliographystyle{rlc}

\appendix

\section{RL Algorithm Supplementary Information}
For completeness, we offer an overview of components of the Oralytics RL algorithm to make ideas in the main paper clear. We again use $i$ to denote the participant and $t$ to denote the decision time. See Appendix A of \citet{nahum2024optimizing}
for full details on the Oralytics RL algorithm.

\subsection{State Space of the RL Algorithm}
\label{app:alg_state}
$S_{i,t} \in \mathbb{R}^d$ represents the state of dimension $d$ (num. of features).

\paragraph{\bf{Baseline and Advantage State Space}}
$f(S_{i,t}) \in \mathbb{R}^5$ denotes the features  used to model both the baseline reward function and the advantage. For Oralytics, the baseline and advantage features are the same (i.e., all $f(S_{i,t})$), but this is a design choice, and they do not have to be.
\begin{enumerate}
    \item \label{alg_state:bias} Bias / Intercept Term $\in \mathbb{R}$
    \item \label{alg_state:tod} Time of Day (Morning/Evening) $\in \{0, 1\}$
    \item \label{alg_state:brushing} $\Bar{B}$: Exponential Average of Brushing Quality Over Past 7 Days (Normalized) $\in \mathbb{R}$
    \item \label{alg_state:a_bar} $\Bar{A}$: Exponential Average of Messages Sent Over Past 7 Days $\in [0, 1]$
    \item \label{alg_state:app} Prior Day App Engagement (Opened App / Not Opened App) $\in \{0, 1\}$
\end{enumerate}

\subsubsection{Modified State Space for Fallback Method}
\label{app:modified_rl_features}
As a fallback method, instead of providing a single action for the current decision time, the Oralytics RL system constructed and provided a full schedule of actions for the entire 70-day study length starting at the current decision time. For decision times within the first 2 weeks from the first day, the algorithm uses the state space as specified above with the following modifications:
\begin{itemize}
    \item Feature~\ref{alg_state:brushing} is replaced with the \textit{Most Recent} Exponential Average Brushing Quality Over Past 7 Days (Normalized) $\in \mathbb{R}$ 
    \item Feature~\ref{alg_state:app} is replaced with the \textit{Best Guess of} Prior Day App Engagement (Opened App / Not Opened App) $= 0$
\end{itemize}

Of the state features presented earlier, features \ref{alg_state:bias}, \ref{alg_state:tod}, and \ref{alg_state:a_bar} were known except for features \ref{alg_state:brushing} ($\bar{B}$) and \ref{alg_state:app} (prior day app engagement). Since future values of $\bar{B}$ were unknown, we imputed this feature with the most recent value of $\bar{B}_{i, t}$ known when the schedule was formed. Namely, the $\bar{B}$ value for decision times $j = t + 2, ...,t + 27$ used  the same $\bar{B}$ value as decision times $t, t + 1$.  For the prior day app engagement feature, we imputed the value to 0, because our best guess was that the participant does not getting a fresh schedule because they did not open the app.

\subsection{Action Space and Policy of the RL Algorithm}
\label{app:action_space}
The algorithm decided between sending a prompt $A_{i, t} = 1$ or not $A_{i, t} = 0$. If $A_{i, t} = 1$, the main controller (described in Section~\ref{sec_system_components}) randomly selected a prompt from the intervention content component. There were 3 different prompt categories: (1) participant winning a gift (direct reciprocity), (2) participant winning a gift for their favorite charity (reciprocity by proxy), and (3) Q\&A for the morning decision time or feedback on prior brushing for the evening decision time. To decide on the exact prompt, the main controller first sampled a category with equal probability and then sampled with replacement a prompt from that category. Due to the large number of prompts in each category, it is highly unlikely that a participant received the same prompt twice. 


We now discuss the policy. Since the RL algorithm in Oralytics was a generalized contextual bandit, the model of the participant environment was a model of the mean reward. To model the reward, the Oralytics RL algorithm used a Bayesian Linear Regression with action centering model: 
\begin{equation}
\label{eqn:blr}
    R_{i, t} = f(S_{i, t})^T \alpha_0 + \pi_{i,t} f(S_{i, t})^T \alpha_1 + (A_{i, t} - \pi_{i, t}) f(S_{i, t})^T \beta + \epsilon_{i,t}
\end{equation}
where $\pi_{i,t}$ is the probability that the RL algorithm selects action $A_{i,t} = 1$ in state $S_{i,t}$. $\epsilon_{i,t} \sim \mathcal{N}(0, \sigma^2)$ and there are priors on $\alpha_{0} \sim \mathcal{N}(\mu_{\alpha_0}, \Sigma_{\alpha_0})$, $\alpha_{1} \sim \mathcal{N}(\mu_{\beta}, \Sigma_{\beta})$, $\beta \sim \mathcal{N}(\mu_{\beta} \Sigma_{\beta})$.

\paragraph{Posterior Updating}
Let $\tau(i,t)$ be the update-time for participant index $i$ that included the current decision time $t$. We needed an additional index because update times $\tau$ and decision times $t$ are not on the same cadence. At update time, the reward model updated the posterior with all history of state, action, and rewards up to that point. Since the RL algorithm used Thompson sampling, the algorithm updated parameters $\mu_{\tau(i,t)}^{\text{post}}$ and $\Sigma_{\tau(i,t)}^{\text{post}}$ of the posterior distribution.

\paragraph{Action-Selection}
The RL algorithm for Oralytics was a modified posterior sampling algorithm called the smooth posterior sampling algorithm. The algorithm selected action $A_{i, t} \sim \text{Bern}(\pi_{i, t})$:

\begin{align}
    \pi_{i,t}
    = \mathbb{E}_{\tilde{\beta} \sim \mathcal{N}(\mu_{\tau(i,t) - 1}^{\text{post}}, \Sigma_{\tau(i,t) - 1}^{\text{post}} )} \left[ \rho(s^\top \tilde{\beta}) \big| \mathcal{H}_{1:n, \tau(i,t) - 1}, S_{i, t} = s \right]
\end{align}
$\tau(i,t) - 1$ was the last update-time for the posterior parameters. Notice that the last expectation above is only over the draw of $\tilde{\beta}$ from the posterior distribution.

In smooth posterior sampling, $\rho$ is a smooth function chosen to enhance the replicability of the randomization probabilities if the study is repeated \citep{zhang2022statistical}. The Oralytics RL algorithm set $\rho$ to be a generalized logistic function:
\begin{equation}
    \rho(x) = L_{\min} + \frac{ L_{\max} - L_{\min} }{ \big[ 1 + c \exp(-b x) \big]^k}
\end{equation}
where asymptotes $L_{\min}, L_{\max}$ are clipping values (bounded away from 0 and 1) to enhance the ability to answer scientific questions with sufficient power \citep{yao2021power}.

\subsection{Reward for the RL Algorithm}
\label{app:reward}
We defined the reward $R_{i, t}$ given to the algorithm to be a function of brushing quality $Q_{i, t}$ (i.e., proximal health outcome) in order to improve the algorithm's learning \citep{trella2023reward}. The reward $R_{i, t}$ was used to update the posterior distribution of the parameters in the reward model described in Appendix~\ref{app:action_space}. The reward for the $i$th participant at decision time $t$ was:
\begin{equation}
\label{reward}
        R_{i, t} := Q_{i, t} - C_{i, t}
\end{equation}

The cost term $C_{i, t}$ allowed the RL algorithm to optimize for immediate healthy brushing behavior, while also considering the delayed effects of the current action on the effectiveness of future actions. The cost term is a function (with parameters $\xi_1, \xi_2$) which took in current state $S_{i, t}$ and action $A_{i, t}$ and outputted the delayed negative effect of currently sending a prompt.

Let $\bar{B}_{i, t}$ and $\bar{A}_{i, t}$ be state features \ref{alg_state:brushing} and \ref{alg_state:a_bar} defined earlier. The cost of sending a prompt was:
\begin{equation}
\label{cost_term}
C_{i, t} := 
\begin{cases}
\xi_1 \mathbb{I}[\bar{B}_{i, t} > b] \mathbb{I}[\bar{A}_{i, t} > a_1] & \\
\hspace{10mm} + \xi_2 \mathbb{I}[\bar{A}_{i, t} > a_2]  & \smash{\raisebox{1.6ex}{if $A_{i, t} = 1$}} \\
0 & \hspace{-0mm} \mathrm{if~} A_{i, t} = 0
\end{cases}
\end{equation}



\section{Additional Forms of Fidelity}
\label{app_other_forms_fidelity}
In this paper, we focus mainly on RL algorithm fidelity; however, there can be other types of fidelity that are important to consider when running an online algorithm in a clinical trial. In addition to algorithm fidelity, there is also \textit{system fidelity} (i.e., how each component in the clinical trial works together) and \textit{participant fidelity} (i.e., participant adherence when participating in the trial). System fidelity can involve verifying correct communication between the sensory-collection device, the data storage system, the computation system where the algorithm runs, and the device that administers the action. Notice that algorithm fidelity is part of system fidelity. On the other hand, participant fidelity can involve concerns such as ensuring the participants are able to correctly download or update the app, keep their smart device charged, actions needed from the participant to obtain sensory data, etc. We make the distinction that issues can arise that impact post-trial analyses and participant experiences that are caused by other components of the system failing (e.g., expired credentials cause participants to lose access to the mobile app), however, this paper focuses on issues that are directly controlled or caused by the RL algorithm.


\end{document}